\documentclass[conference,compsoc]{IEEEtran}
\ifCLASSOPTIONcompsoc
  \usepackage[nocompress]{cite}
\else
  \usepackage{cite}
\fi
\usepackage{comment}
\usepackage{graphicx} 
\usepackage{amssymb}
\usepackage{booktabs}
\usepackage{etoolbox}
\makeatletter
\patchcmd{\@makecaption}
  {\scshape}
  {}
  {}
  {}
\makeatother

\ifCLASSINFOpdf
\else
\fi
\usepackage{amsmath}
\DeclareMathOperator*{\argmax}{arg\,max}

\begin{document}
%
\title{GlobalWalk: Learning Global-aware Node Embeddings via Biased Sampling}

\author{\IEEEauthorblockN{Zhengrong Xue}
\IEEEauthorblockA{Shanghai Jiao Tong University\\
Shanghai, China\\
Email: xuezhengrong@sjtu.edu.cn}
\and
\IEEEauthorblockN{Zi'ao Guo}
\IEEEauthorblockA{Shanghai Jiao Tong University\\
Shanghai, China\\
Email: 1115712590@sjtu.edu.cn}
\and
\IEEEauthorblockN{Yiwei Guo}
\IEEEauthorblockA{Shanghai Jiao Tong University\\
Shanghai, China\\
Email: cantabile\_kwok@sjtu.edu.cn}}


%


\maketitle

\begin{abstract}
Popular node embedding methods such as DeepWalk follow the paradigm of performing random walks on the graph, and then requiring each node to be proximate to those appearing along with it. Though proved to be successful in various tasks, this paradigm reduces a graph with topology to a set of sequential sentences, thus omitting global information. To produce global-aware node embeddings, we propose GlobalWalk, a biased random walk strategy that favors nodes with similar semantics. Empirical evidence suggests GlobalWalk can generally enhance embedding's global awareness.
\end{abstract}


%
\IEEEpeerreviewmaketitle

\section{Introduction}

Many would agree that we are, as the saying goes, defined by the people we surround ourselves with. When discussing the essence of human nature, Karl Marx conveys similar ideas in a more academic way, ``we are the ensemble of the social relations". From the view of researchers in computer science, Marx's theory on human nature can be described as, every node (atom individual) in a graph (the social web) can be represented by its neighbors (people around). This, in fact, is exactly the philosophy behind many node embedding approaches.

Node embedding maps a node from the sparse and high-dimensional graph space to a dense and low-dimensional vector space. If the embedding is appropriately learned, it can provide much convenience for the utilization of other machine learning algorithms and help boost the performance of downstream tasks~\cite{cai2018comprehensive}. After the pioneering work of DeepWalk~\cite{perozzi2014deepwalk}, a popular paradigm nowadays is to convert a graph into a set of ``sentences of nodes" by taking random walks on the graph~\cite{tang2015line,grover2016node2vec,ribeiro2017struc2vec}, and then feed the set to word2vec~\cite{mikolov2013efficient}.

While these methods have exhibited tremendous success in tasks such as link prediction and community detection, there is undeniably a crucial difference between a graph and a paragraph. A paragraph consists of sequential sentences, which is in born aligned with word2vec in structure. In contrast, the structure of a graph is topological. The set of ``sentences of nodes" may be a good descriptor for low-order proximity, thus well approximating local information of a graph. Nonetheless, it is less likely to be aware global information of topology on a graph (e.g. communities), which is essentially what tells graphs apart from paragraphs.

\begin{figure}[t]
\centering
\includegraphics[width=3.3in]{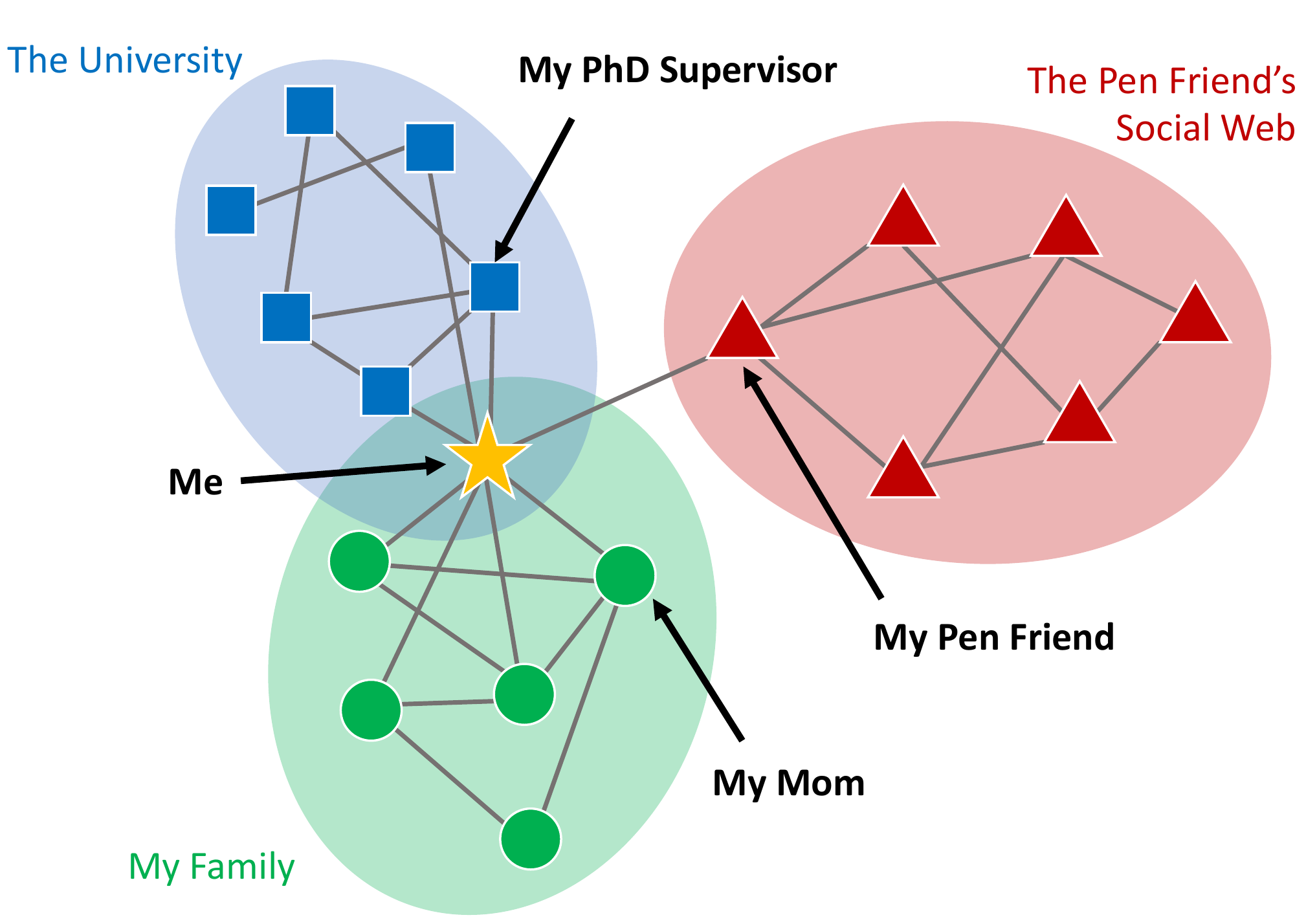}
\caption{Based on Marx's theory on human nature, suppose ``I" am jointly defined by my mom, my PhD supervisor, and my pen friend. Apparently, my mom and my PhD supervisor contribute more to the definition of myself than my pen friend. This is not only due to closer connections between me and them (local information). More importantly, it is also because both my supervisor and I belong to the community of the university, and my mom and I both belong to the community of my family (global information).}
\label{motivation}
\end{figure}

Recent advances in node embedding start to take notice of the importance of global information, and propose various techniques to make use of it. The majority of them  explicitly leverage community information to constrain the optimization process of node embeddings. For example, ComE~\cite{cavallari2017learning} generates GMM to depict communities and requires node embeddings to fit the GMM distribution. In CNRL~\cite{tu2018unified}, communities are generated from Latent Dirichlet Allocation, and the average of all embeddings in a sentence is considered as the community embedding. While these methods do improve the performance under certain scenarios, we find them very sensitive to specially designed architectures of community descriptors as well as the trade-off factor between local and global information, which may hinder them from broader applications.

In view of the drawbacks in existing techniques, we propose GlobalWalk, a prior-free and easy-to-implement method for learning global-aware node embeddings. GlobalWalk adds global information to the paradigm of ``sentences of nodes" by performing a biased random walk. Unlike current approaches that consider biased sampling on a graph-structural level, still stuck in local information, we do biased sampling on a semantic level, which can grasp global information.

Our motivation comes from a natural extension to Marx's theory on human nature: different people (neighbors) contribute differently to the definition of a person (node). Please see Figure~\ref{motivation} for more explanation. Borrowing some concepts from mathematics and may abusing them a little, we claim one person (node) is not a ``arithmetic average" of people around him (neighbors), but a ``weighted average" of them. Therefore, when performing random walks on a graph, we should always actively favor those with more consistent semantics (e.g. belonging to the same community), but pay less attention to the distant ones. In practice, the distance between node embeddings in the latent space is used to scale how semantically consistent two nodes are, and a simulated annealing procedure is taken to balance local and global information. 

We implement GlobalWalk based on the backbone of DeepWalk, and evaluate its performance on several datasets. Despite no evident improvements on the task of link prediction that heavily relies on local information, GlobalWalk is found to perform better on the task of community detection, which requires more global information. Besides, ablation studies and visualization results are also presented to further demonstrate the effectiveness of the proposed GlobalWalk approach.


\section{Related Work}

\subsection{Node Embedding on Graph}

It is very intuitive to put natural languages and graphs in an analogy --- a node in a graph can be thought of as a word, and a ``bigram" much resembles an edge.
DeepWalk\cite{perozzi2014deepwalk} is the first to utilize this idea together with the technique of embedding popularized by word2vec\cite{mikolov2013efficient}, to learn low-dimensional representation for every node in a graph.
By sampling fixed-length sequences in graphs,
DeepWalk uses neural networks to optimize the conditional probability of a node appearing within its context, in a way much similar with Skip-Gram\cite{skipgram}.

LINE\cite{tang2015line} extends DeepWalk with {first-order proximity} between nodes, while DeepWalk only considers second-order proximity.
By optimizing first-order proximity, LINE produces embeddings that are closer  when two nodes are directly adjacent.
This method is applicable to all types of graphs, while DeepWalk can only be performed on unweighted graphs.

Based on the former methods, node2vec\cite{grover2016node2vec} discusses how to obtain ``sentences" of nodes that are more meaningful.
node2vec claims that the sampled sequences must both embody homophily (i.e. close nodes should have close embeddings) and structural equivalence (i.e. nodes that share similar structures should also have similar embeddings, even if not adjacent).
It achieves this by mixing breadth-first and depth-first search.

These methods are popular and foundational researches that integrate deep representation learning into graphs by focusing on local information conveyed in random walks.
Such technique is further applied to a wide variety of tasks, such as node classification and clustering\cite{cao2015grarep,tian2014learning}, link prediction\cite{ou2016asymmetric}, graph visualization\cite{wang2016structural} and more.
A branch of later works pays attention to global information in graphs, which we summarize as the ``global-aware" node embedding methods.

\subsection{Global-aware Node Embedding}

Community-enhanced representation learning (CNRL)\cite{tu2018unified}
simultaneously
detects community distribution of each vertex and learns embeddings of both vertices and communities.
Analogizing communities with ``topics" in NLP,
CNRL uses Gibbs sampling as in LDA\cite{LDA} for community assignment, and then typical node2vec techniques for representation learning.
Thus it learns better representation for community-related tasks.

ComE\cite{cavallari2017learning} takes a different perspective.
It introduces community embeddings 
to optimize the node embedding by introducing a community-aware high-order proximity.
It extends LINE with a multivariate Gaussian mixture distribution over node embeddings.
This GMM serves as community distribution to optimize the high-order node embedding proximity.
An iterative process is taken to both optimize the node embeddings and the community distribution.

vGraph\cite{vGraph} also learns community membership and node representation collaboratively.
It designs a variational inference algorithm to regularize the community membership of neighboring nodes to be similar in latent space.
Unlike previous methods, 
vGraph is highly general and can apply to detect hierarchical communities by a $d$-level tree.

Similar global-aware embedding methods include\cite{wang2017community,cao2015grarep,ou2016asymmetric}.
Though these methods indeed improves performance on community-related downstream tasks,
they highly depend on a manually defined concept of ``community".
These community assignments are often ambiguous and are not the natural representation of global information.
Thus they are either overly complicated or sensitive to tasks, which limits their applicability.

\section{Background}

\subsection{Problem Formulation}
A real-life network can be represented by a graph $G=(V,E)$, where $V$ is the set of nodes and $E$ is the set of edges. The property of the network decides the graph is either directed or undirected. A node embedding of a graph $G=(V,E)$ is a mapping from the sparse and high-dimensional node space to a dense and low-dimensional embedding space, $\Phi:V\to\mathbb{R}^d$. We expect the node embedding to be well aware of both local and global information, thus preserving the topology of the graph.

\subsection{DeepWalk and node2vec}
The pioneering work of DeepWalk provides a popular framework for node embedding problems, whose core idea is to approximate a complex graph via a stream of simple, short random walks on the graph. The procedure of DeepWalk can be summarized into two steps. The first step is to perform random walks on the graph. Every walk is considered as a ``sentence of nodes". The second step of DeepWalk is to optimize the generated walks by following the Skip-Gram model in word2vec. Intuitively, the Skip-Gram model requires the embeddings of context nodes to be similar to that of the target node $u$
\begin{align}
\label{skip_gram}
    \argmax_\Phi\sum_{u\in V}\log p(C(u)|u),
\end{align}
where $C(u)$ denotes the set of ``neighbors" of node $u$. Note that ``neighbors" here are not necessarily immediate neighbors of the node, but also include those appearing in the generated walks. The Skip-Gram model assumes context nodes (words) to be independent of each other. Thus, Equation~\ref{skip_gram} can be rewritten as
\begin{align}
    \argmax_\Phi\sum_{u\in V}\sum_{v\in C(u)}\log p(v|u),
\end{align}
where $p(v|u)$ is defined by a softmax function
\begin{align}
    p(v|u)=\frac{\exp({\Phi(v)^\top\Phi(u)})}{\sum_{w\in V}\exp({\Phi(w)^\top\Phi(u)})}.
\end{align}

Unlike DeepWalk which takes a fair random walk, node2vec takes structure similarity into account and performs a biased random walk
\begin{align}
    p(c_i=x|c_{i-1}=v)\propto\alpha_{pq}(t,x),
\end{align}
where $c_i$ denotes the $i$th node in a walk, $t$ is the last visited node before $v$, and $\alpha_{pq}(t,x)$ controls the random walk behaves more like a depth-first search or a breadth-first search
\begin{equation}
    \alpha_{pq}(t,x)=
    \begin{cases}
    1/p &\text{ if } d_{tx}=0 \\
    1 &\text{ if } d_{tx}=1 \\
    1/q &\text{ if } d_{tx}=2
    \end{cases}.
\end{equation}

Similar to node2vec, our method performs a biased random walk as well. Nevertheless, the nature that node2vec only cares about second-order proximity makes it inevitably indulged in local information. In comparison, our method leverages semantics of nodes to guide the bias, thus implicitly considering the global information.

\section{Proposed Methods}
\label{section_GlobalWalk}
\subsection{GlobalWalk}
The ultimate goal of GlobalWalk is to produce high-quality node embeddings that coherently integrate local and global information. Classic approaches such as DeepWalk and node2vec are not aware of global information at all. Recent advances typically utilize the prior that global information can be represented with generated community distributions, and explicitly force node embeddings to follow those distributions. However, the concept of community itself is ambiguously defined. What's worse, the number of communities is often required to be predefined, which is not practical in many actual cases. Thus, we don't think communities to be the best indicator for global topology of a graph.

Unlike all the existing approaches, the proposed GlobalWalk takes the perspective of semantics. We have noticed that even with very similar structures, different neighbors of a node can have dramatically distinct impact on it. The intrinsic reason behind is that nodes with close semantics naturally tend to have links with each other. We call those links a ``necessity". In comparison, those with distant semantics may also have links, but definitely with a smaller probability. We call those links a ``coincidence". In the example of Figure~\ref{motivation}, it is a coincidence that ``I" get acquainted with the pen friend, but it is a necessity that ``I" have strong connections with my mom.

It is fairly intuitive that events of necessity should appear more frequently than those of coincidence. To make it happen, GlobalWalk takes advantage of the node embeddings in latent space to scale the semantics between two nodes. When performing a random walk that currently stands at a node $u$, GlobalWalk will calculate the Euclidean distance between the embeddings of $u$ and all its neighbors. The likelihood that the walk takes a neighbor $v$ is in negative correlation with the distance between the embeddings of $u$ and $v$.

Reasonable as it looks like, there are two crucial problems that we have to dive into. Firstly, how should the likelihood function be defined so that it can suit the graph settings? Secondly, it seems that the embeddings that are used to scale semantics are at the same time the output of our method. How could we avoid getting stuck into the chicken-and-egg problem?

For the first problem, we provide three options --- a inversely proportional function, a threshold function, and a shifted exponential function. They will introduce them in Section~\ref{sec_distance}. For the second problem, we propose a simulated annealing procedure, will be discussed in Section~\ref{sec_anneal}. In brief, GlobalWalk behaves just like DeepWalk at first so that the training process is enabled, and then gradually takes semantics into consideration so that it can be aware of global information.

\subsection{Likelihood Functions}
\label{sec_distance}
In this section, we propose three different kinds of likelihood functions. Each function takes in a node $u$ and one of its neighbors $v$, and returns the likelihood that the biased random walk choose $v$. As has been stated, all of the functions should be in negative correlation with the distance of embeddings $\xi(u,v)$. The distance here refers to the normalized 2-norm distance of the embeddings on Euclidean space, which is given by
\begin{align}
    \xi(u,v)=\frac{\left \| \Phi(u)-\Phi(v) \right \|_2 }{\max_{w\in\mathcal{N}(u)}\left \| \Phi(u)-\Phi(w) \right \|_2},
\end{align}
where $\mathcal{N}(u)$ denotes the set of neighbors of $u$.

The first likehood function is a inversely proportional function
\begin{equation}
    p_{\text{inv}}(v|u)=
    \begin{cases}
    1/\xi(u,v) &\text{ if } \xi(u,v) > \epsilon_{\text{inv}}\\
    1/\epsilon_{\text{inv}} &\text{ if } 0 <= \xi(u,v) <= \epsilon_{\text{inv}}\\
    \end{cases},
\end{equation}
where $\epsilon_{\text{inv}}$ is a small threshold set to be $0.01$ here to prevent division by zero error. The second function is a threshold function
\begin{equation}
    p_{\text{thr}}(v|u)=
    \begin{cases}
    \epsilon_{\text{thr}} &\text{ if } \xi(u,v) > \epsilon_{\text{thr}} \\
    1/\epsilon_{\text{thr}} &\text{ if } 0 <= \xi(u,v) <= \epsilon_{\text{thr}}\\
    \end{cases},
\end{equation}
where $\epsilon_{\text{thr}}$ is a threshold set to be $0.5$. The third function is a shifted exponential function
\begin{equation}
    p_{\text{exp}}(v|u)= c - \exp(\xi(u,v)),
\end{equation}
where $c$ is a constant set to be $2$.

Intuitively, we would like to punish those too far away from the target node in semantics, but keep relatively fair to the other neighbors. In this sense, the threshold function and the shifted exponential function, which can be regarded as a soft version of the threshold, are better choices. Empirically, we find that the shifted exponential function generally attains the best performance on most tasks and datasets.

\subsection{Simulated Annealing}
\label{sec_anneal}
Different from other methods, node embedding serves a dual role in GlobalWalk. On one hand, it is the final result we want to acquire. On the other hand, it should provide global information for the training process. To conquer the tough problem of chicken-and-egg, we adopt a simulated annealing procedure.

In the first epoch, we cannot count on the untrained embedding to give anything useful. Therefore, we do exactly the same thing as DeepWalk. In this case, the random walk takes one neighbor with completely likelihood
\begin{align}
\label{p0}
    p_0(v|u)=1/|\mathcal{N}(u)|.
\end{align}

Starting from the second epoch, global information are gradually added to the decision process. With the training process going and the node embeddings becoming more trustworthy, we tend to give more weight to global information
\begin{align}
\label{pt}
    p_t(v|u)=(1-t\beta)/|\mathcal{N}(u)|+t\beta p_{\text{bias}}(v|u),
\end{align}
where $t$ is the index of epochs, $\beta$ is the annealing factor which controls how aggressively the simulated annealing goes, and $p_{\text{bias}}$ is a likelihood function selected from the set $\{p_{\text{inv}}, p_{\text{thr}}, p_{\text{exp}}\}$. Note that Equation~\ref{p0} is actually a special case of Equation~\ref{pt} when $t=0$.

\section{Experiments}
As GlobalWalk takes both global and local features into consideration, we evaluate it on community detection and link prediction, which gives more attention to global and local information, respectively.
\subsection{Settings}

\noindent \textbf{Datasets.}
For community detection, we evaluate GlobalWalk on $4$ datasets, including email-Eu-core, Football, PoliticalBlogs and Politicalbooks. For link prediction, we use $2$ datasets, ca-AstroPh and email-Eu-core. Statistics of datasets are provided in TABLE \ref{dataset}. Limited by time and computational resources, we conduct experiments on smaller-scale datasets. To include as many settings as possible, the datasets cover both sparse and dense graphs, directed and undirected graphs, and binary and multi-classification, which we believe are sufficient to demonstrate the effectiveness of GlobalWalk.

\begin{table}[ht]
    \caption{Dataset statistics. $|V|$: number of nodes, $|E|$: number of edges, CN: number of communities, ED: direction of edges, CD: community detection, LP: link prediction.}
    \centering
    \begin{tabular}{cccccc}
        \toprule
        Dataset & $|V|$ & $|E|$ & CN & ED & Task  \\
        \midrule
        Politicalbooks  & 105   & 441  & 3 & Undirected & CD  \\
        Football  & 115   & 613  & 12 & Undirected & CD  \\
        PoliticalBlogs  & 1,490  & 19,090   & 2 & Directed & CD  \\
        email-Eu-core  & 1,005   & 25,571   & 42 & Directed & LP \& CD  \\
        ca-AstroPh  & 18,772   & 198,110   & / & Undirected & LP  \\
        \bottomrule       
    \end{tabular}
    \label{dataset}
\end{table}

\noindent \textbf{Evaluation Metrics.}
For community detection, we use \textit{ACC} to measure the performance of the algorithms, which is defined in Equation \ref{ACC}. $N$ is the total number of nodes and $N_{true}$ refers to the number of nodes that are correctly classified.
\begin{equation}
    ACC = \frac{N_{true}}{N}.
    \label{ACC}
\end{equation}
For link prediction, we use \textit{AUC} as Equation \ref{AUC} defines. $\mathbf{D^0}$ is the set of node pairs without edges, while $\mathbf{D^1}$ is the set of those with edges. $\mathbf{1}$ is an indicator. $f$ represents the probability that an edge exists between the node pair. We choose an arbitrary function inversely proportional to the distance of the two nodes for implementation.
\begin{equation}
    AUC = \frac{\Sigma_{t_0 \in \mathbf{D^0}} \Sigma_{t_1 \in \mathbf{D^1}} \mathbf{1}[f(t_0) < f(t_1)]}{|\mathbf{D^0}||\mathbf{D^1}|}.
    \label{AUC}
\end{equation}

\noindent \textbf{Comparative Methods and Hyperparameters.}
DeepWalk\cite{perozzi2014deepwalk} and node2vec\cite{grover2016node2vec} are selected as the baselines. Since both node2vec and our GlobalWalk have the backbone of DeepWalk, they share a considerate number of parameters. In our experiments, the length of a path generated by random walk is set to $80$, and each node generates $10$ paths in each epoch. We train the embeddings of nodes for $3$ epochs in total. The window size of word2vec is $10$. Each node is embedded to a $64$-dimensional vector. Besides, K-Means is used to cluster nodes into communities for community detection. In node2vec, a larger $p$ indicates it tends to explore, while a larger $q$ indicates it is more like to wander. Thus, we set $p=2, q=0.5$ for community detection, and $p=0.5, q=2$ for link prediction. In GlobalWalk, we choose $p_{\text{exp}}$ as the likelihood function, and set $\beta=0.2$. The results of other likelihood functions and $\beta$'s are presented in ablation studies in Section~\ref{AS}.

\subsection{Experimental Results}
\begin{figure}[t]
    \centering
    \includegraphics[width=3.3in]{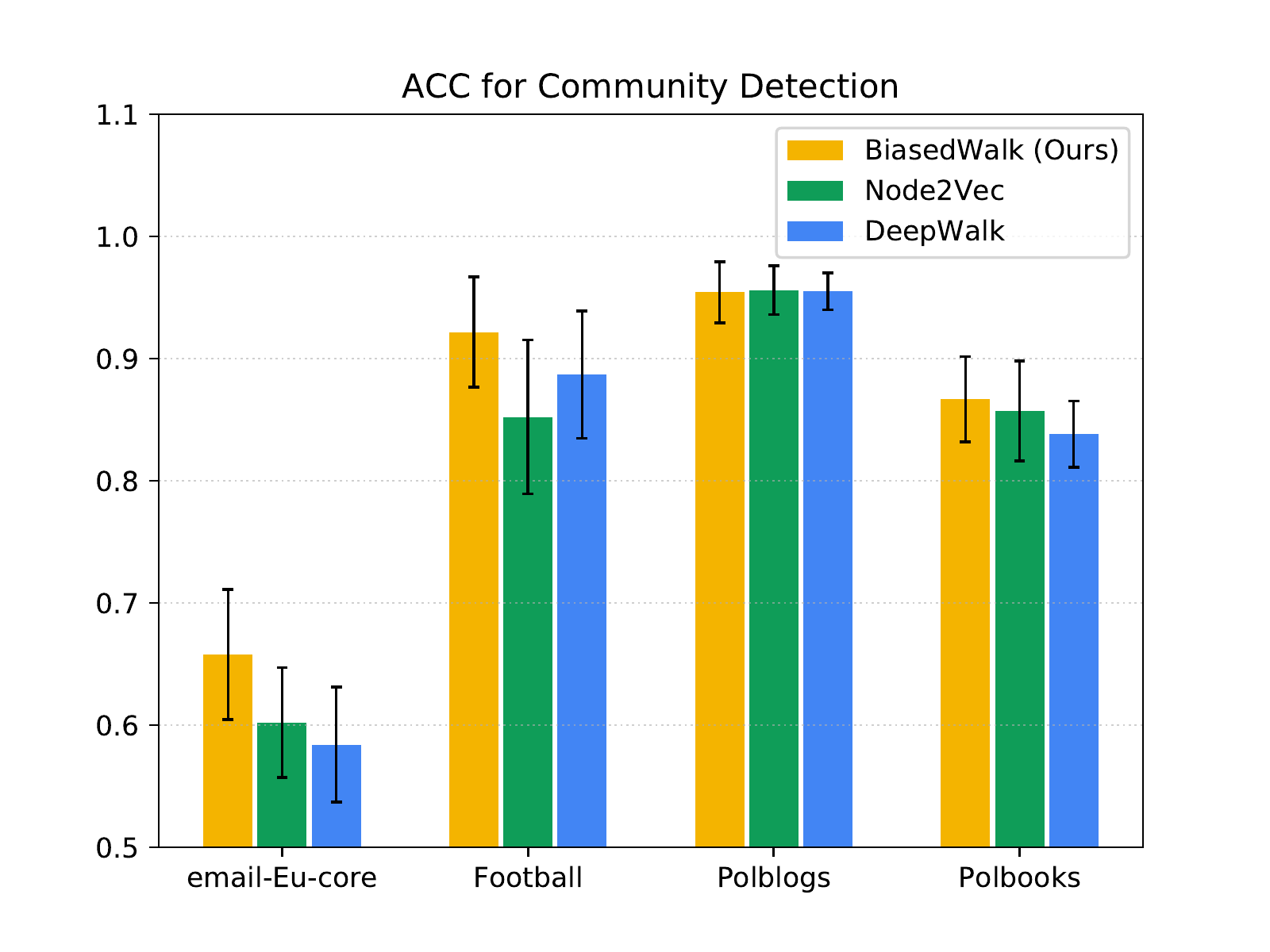}
    \caption{Evaluation (in terms of \textit{ACC}) on community detection datasets}
    \label{fig:cd_result}
\end{figure}

Figure \ref{fig:cd_result} shows the results on community detection. GlobalWalk outperforms baseline methods on 3 out of 4 datasets in terms of \textit{ACC}, and shows identical performance on the remaining 1 dataset, which is a simple binary classification. This supports our claim that GlobalWalk can better grasp global information by considering the semantics of nodes.

\begin{figure}[t]
    \centering
    \includegraphics[width=3.3in]{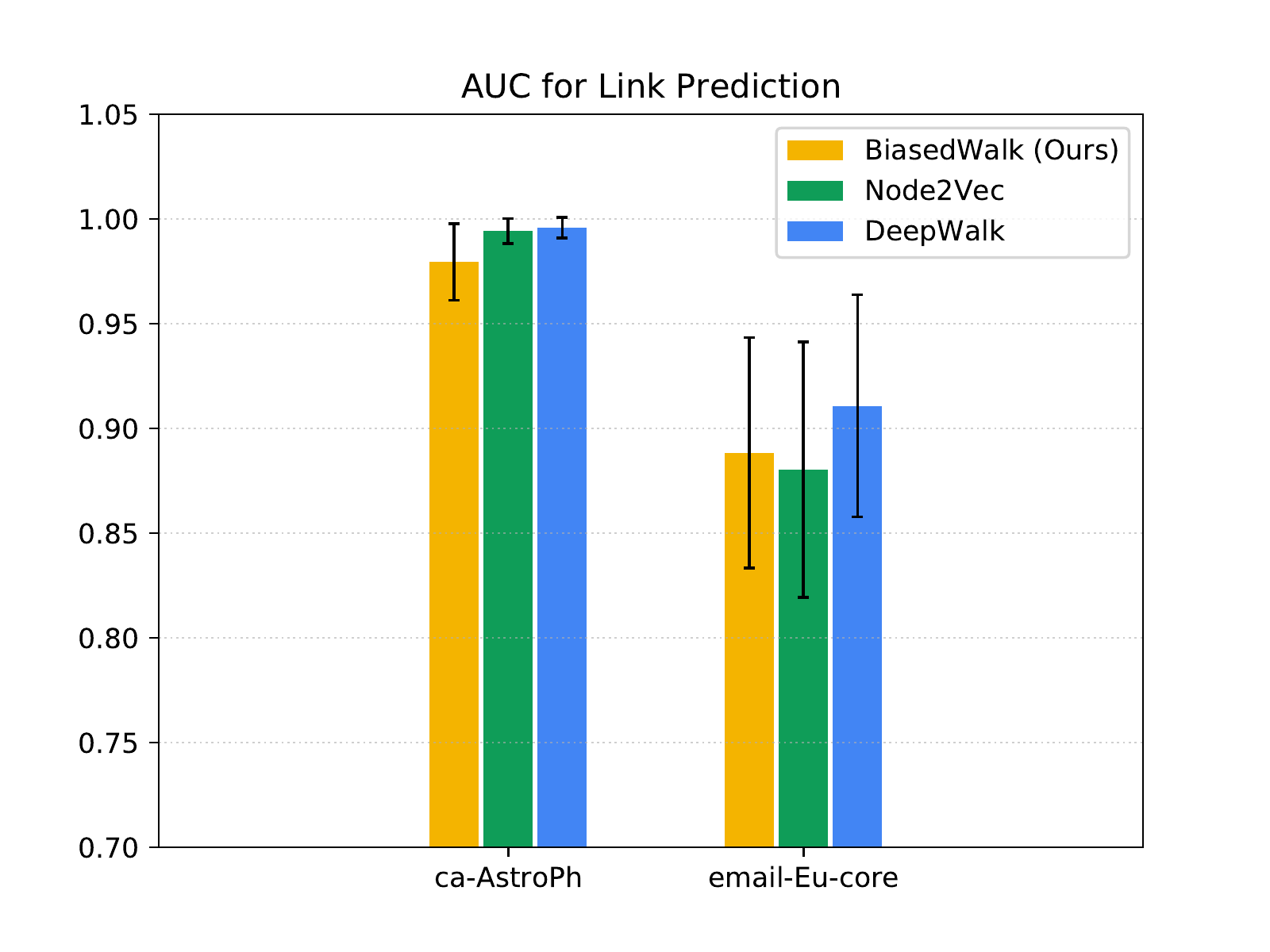}
    \caption{Evaluation (in terms of \textit{AUC}) on link prediction datasets}
    \label{fig:lp_result}
\end{figure}

The results on link prediction are shown in Figure \ref{fig:lp_result}. Compared to DeepWalk and node2vec, GlobalWalk exhibits comparable performance. It is predictable in that the design of GlobalWalk makes it not good at link prediction, as there is a trade-off between local and global features. Nevertheless, by adjusting $\beta$, we can still get competitive performance on link prediction.

\subsection{Ablation Studies} \label{AS}
We conduct ablation studies on the task of community detection on the email-Eu-core dataset. We apply different likelihood functions and $\beta$'s to evaluate how the performance of GlobalWalk changes with hyperparameters. The results are shown in TABLE~\ref{ablation}. 

$p_{\text{exp}}$ achieves the best \textit{ACC} score among all the likelihood functions, which is in line with our intuition. As for $\beta$, it can be observed that when $\beta$ is set to be too high or too low, the performance degrades. This is because the process of annealing should be maintained at a moderate rate to acquire better convergence. Comprehensively speaking, $\beta=0.2$ and $p_{\text{exp}}$ for likelihood function attains the best performance for the task of community detection.

\begin{table}[ht]
    \caption{Ablation study results on community detection task on email-Eu-core dataset.}
    \centering
    \begin{tabular}{cc|cc}
        \hline
        Likelihood Function & \textit{ACC} & Annealing Fctor & \textit{ACC}  \\
        \hline
        $p_{\text{inv}}$  & 0.6239    & 0.3   & 0.6470  \\
        $p_{\text{thr}}$  & 0.6129   & 0.2  & \textbf{0.6509}  \\
        $p_{\text{exp}}$  & \textbf{0.6577}& 0.1 & 0.6259  \\
        \hline
    \end{tabular}
    \label{ablation}
\end{table}

\subsection{Visualization of Community Detection}
\begin{figure}[ht]
    \centering
    \includegraphics[width=3.0in]{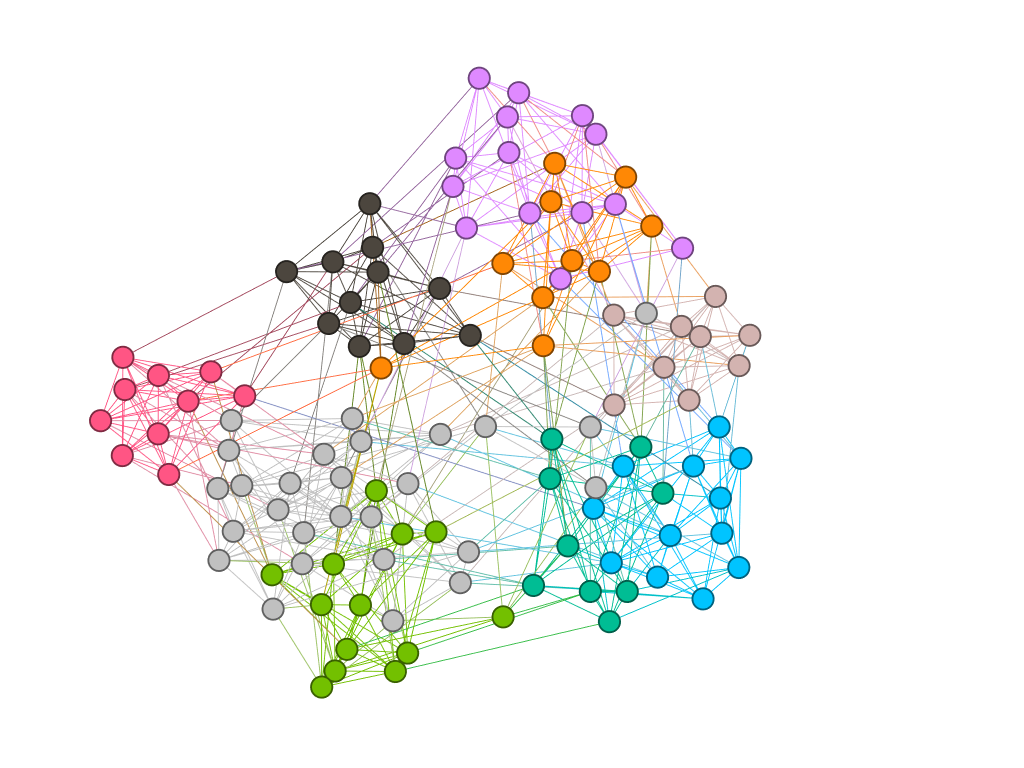}
    \caption{Visualization of communities detected by GlobalWalk on Football dataset.}
    \label{fig:visual}
\end{figure}

Visualization of predicted communities on Football dataset is shown in Figure \ref{fig:visual}, where nodes belonging to the same community are assigned to the same color. It can be observed that most communities are in a compact structure, which indicates that our GlobalWalk performs well on this task.

\bibliographystyle{IEEEtran}
\bibliography{cit}
%

\end{document}